\documentclass{ieeeaccess}
\usepackage{cite}
\usepackage{amsmath,amssymb,amsfonts}
\usepackage{algorithmic}
\usepackage{graphicx}
\usepackage{textcomp}

\usepackage{hyperref} 
\usepackage{booktabs} 
\usepackage{arydshln} 
\usepackage{xcolor} 
\usepackage{colortbl} 
\usepackage{soul} 
\usepackage{graphbox} 

\newcommand*{\dittoclosing}{--- \raisebox{-0.5ex}{''} ---}

\makeatletter
\AtBeginDocument{\DeclareMathVersion{bold}
\SetSymbolFont{operators}{bold}{T1}{times}{b}{n}
\SetSymbolFont{NewLetters}{bold}{T1}{times}{b}{it}
\SetMathAlphabet{\mathrm}{bold}{T1}{times}{b}{n}
\SetMathAlphabet{\mathit}{bold}{T1}{times}{b}{it}
\SetMathAlphabet{\mathbf}{bold}{T1}{times}{b}{n}
\SetMathAlphabet{\mathtt}{bold}{OT1}{pcr}{b}{n}
\SetSymbolFont{symbols}{bold}{OMS}{cmsy}{b}{n}
\renewcommand\boldmath{\@nomath\boldmath\mathversion{bold}}}
\makeatother

\def\BibTeX{{\rm B\kern-.05em{\sc i\kern-.025em b}\kern-.08em
    T\kern-.1667em\lower.7ex\hbox{E}\kern-.125emX}}

\begin{document}
\history{Date of publication: 16 October 2026. The final published version is available at http://doi.org/10.1109/ACCESS.2026.3744200.}
\doi{10.1109/ACCESS.2026.3744200}

\title{LLaVA-CKD: Bottom-Up Cascaded Knowledge Distillation for Vision-Language Models}
\author{\uppercase{Nikolaos Gkalelis}\href{https://orcid.org/0000-0001-6741-3334}{{\includegraphics[width=0.12in]{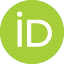}}}\authorrefmark{1}, 
\IEEEmembership{Member, IEEE}, AND \uppercase{Vasileios Mezaris}\href{https://orcid.org/0000-0002-0121-4364}{{\includegraphics[width=0.12in]{images/orcidicon.png}}}\authorrefmark{1}, \IEEEmembership{Senior Member, IEEE}}
\address[1]{Centre for Research and Technology Hellas (CERTH), Information Technologies Institute (ITI), 57001 Thessaloniki, Greece.}
\tfootnote{This work was supported by the European Union (EU) Horizon Europe Program under Grant 101214398 ELLIOT.\\ The publication of the article in OA mode was financially supported by HEAL-Link.}
\markboth
{Gkalelis N. \headeretal: LLaVA-CKD: Bottom-Up Cascaded Knowledge Distillation for Vision-Language Models}
{Gkalelis N. \headeretal: LLaVA-CKD: Bottom-Up Cascaded Knowledge Distillation for Vision-Language Models}
\corresp{Corresponding authors: Nikolaos Gkalelis (e-mail: gkalelis@iti.gr) and Vasileios Mezaris (e-mail: bmezaris@iti.gr)}

\begin{abstract}
Large Vision-Language Models (VLMs) are successful in addressing a multitude of vision-language understanding tasks, such as Visual Question Answering (VQA), but their memory and compute requirements remain a concern for practical deployment. A promising class of techniques for mitigating this concern is Knowledge Distillation, where knowledge from a high-capacity Teacher network is transferred to a considerably smaller Student network. However, the capacity gap between the two networks is both a blessing and a curse: the smaller the Student network, the better its efficiency, and the stronger (typically, also larger) the Teacher, the more knowledge it carries; yet, beyond a point, the larger capacity gap between the two leads to worse knowledge transfer. To counter this effect, we propose a bottom-up cascaded knowledge distillation (CKD) framework. Instead of treating knowledge transfer as an activity involving one high-capacity Teacher (or an ensemble of such), inspired by human formal education systems, we introduce one (potentially, more) additional Teacher(s) of intermediate capacity that gradually bring the Student network to the next level, where the next (higher-capacity) Teacher can take over. We provide a theoretical analysis to motivate why progressive reduction of the Teacher-Student gap may improve distillation. 
We apply the proposed framework on models built upon the LLaVA methodology and evaluate the derived models on seven standard, publicly available VQA benchmarks, demonstrating competitive performance with recent compact VLMs. Source code for cascaded distillation and trained models are available at \url{https://github.com/IDT-ITI/LLaVA-CKD}.
\end{abstract}

\begin{keywords}
Vision Language Models, Knowledge Distillation, LLaVA, Capacity Gap, Deep Learning.
\end{keywords}

\titlepgskip=-15pt

\maketitle

\section{Introduction}
\label{S:Intro}

\PARstart{V}{ision}-Language Models (VLMs) have demonstrated remarkable performance in a wide range of application domains, ranging from fundamental tasks, such as image captioning, to more advanced ones such as scientific discovery and embodied robot navigation in outdoor environments   \cite{KotsiantisVlmlReviewIeeeAccess2026,MllmReviewLiang2024,visintll_Xie2025}.
However, the size and substantial inference cost of SotA models, attributed to scaling laws for the development of high-performance VLMs \cite{ScalingLawsKaplan2020,TrainCompOptHoffmannNips2022}, impact their widespread deployment, especially in limited computational resource environments \cite{efficient_Jin2025,VlmEfficientEdgeSharshar2025}.

\begin{figure*}[tbp]
\centering
\includegraphics[width=7in]{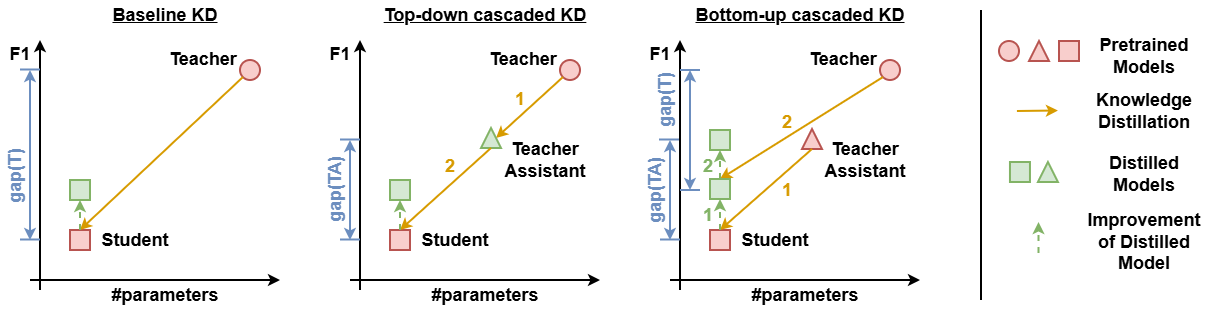}
\caption{Illustration of different KD strategies. Baseline KD suffers from the large capacity gap between the Student and the Teacher. Top-down cascaded KD attempts to reduce the capacity gap by introducing a Teacher Assistant (TA), to be used in place of the larger Teacher model. In contrast, the proposed bottom-up cascaded KD leverages both the TA and the original Teacher model for directly transferring knowledge to the Student, while gradually closing the gap between the latter two.
}
\label{fig:LlavaCkdConcept}
\end{figure*}

A straightforward strategy for creating compact VLMs is to scale-down their larger counterparts, i.e., simply train smaller network architectures.
However, this typically leads to significant performance drop.
Other popular strategies include quantization, pruning, and Knowledge Distillation (KD); their advantage being that they can reduce the size of large-scale VLMs while preserving a good part of the knowledge carried by these larger models. Among the above, KD, a well-established approach to transfer knowledge from a high-capacity Teacher to a lightweight Student network, is recently getting increasing attention in relation to VLMs, as it can easily complement other approaches (e.g., quantization) for the creation of even more compact but well-performing models \cite{ZhangDistilReviewIeeeAccess2025,KotsiantisVlmlReviewIeeeAccess2026}.
KD techniques for VLMs, e.g., \cite{LlavaKdICCV2025}, utilize suitable losses for effectively processing the multimodal knowledge of the Teacher, and a multi-step distillation strategy with appropriately curated instruction datasets to gradually transfer knowledge from the Teacher to the Student. However, KD approaches are known to face difficulties when the Teacher is much stronger than the Student, a phenomenon known as the ``capacity gap'' in distillation \cite{BusbridgeDistLawsIcml2025}.

A rather small body of literature has attempted to alleviate the capacity gap in distillation, and most relevant works focus on previous-generation network architectures (e.g. CNNs) \cite{TeachAssistAaai2020,DenseGuideTaIccv2021,TaSentimAnalysisDongSsci2023}. In the above works, a new network is introduced, whose capacity is somewhere in between the Teacher's and the Student's (termed ``Teacher Assistant (TA)''), and distillation follows a top-down strategy comprising of two stages (Fig. \ref{fig:LlavaCkdConcept}, middle diagram): 1) train the TA via KD from the Teacher, and, 2) perform KD from the distilled TA to the Student.
This reduces the capacity gap between the TA and the Student (compared to the gap between the original strong Teacher and the Student) but leads to not fully exploiting the knowledge carried by the strong Teacher. Contrarily to such previous approaches, we do not use the strong Teacher in a top-down approach, i.e., to create better TA(s) that will ultimately lead to a single TA being used for transferring knowledge to the Student.
Instead, we propose a bottom-up approach (Fig. \ref{fig:LlavaCkdConcept}, right diagram), where the TA is used first for transferring knowledge to the Student, followed by the strong Teacher also transferring knowledge to the improved Student that now exhibits a smaller capacity gap with its Teacher.  

In summary, our major contributions are:
\begin{itemize}
    \item We present the first, to the best of our knowledge, bottom-up cascaded knowledge distillation approach for VLMs, to gradually improve the performance of small-scale Student VLMs using more than one pretrained Teachers -- specifically, a Teacher and a TA -- of varying strength.
    \item We provide both experimental evidence on seven benchmarks, to document how the proposed approach works, and theoretical insights to motivate bottom-up cascaded distillation.
\end{itemize}

\section{Related Work}
\label{S:RelWork}

In the following, we discuss recent techniques for the creation of small-scale VLMs.
For a more comprehensive survey, the reader is referred to \cite{efficient_Jin2025} and \cite{VlmEfficientEdgeSharshar2025}.

\textbf{Compact models.}
A straightforward approach to create a small-scale VLM is simply
scaling down large-scale ones and using appropriate training datasets for training them. MobileVLM v1 \cite{mobVLM2023} is the first open VLM built using publicly available datasets that is customized for resource-constrained environments. 
MobileVLM v2 \cite{mobVLMv2_2024} improves upon its v1, by enhancing the vision-language connector, adopting the LLaVA training strategy \cite{LiuLLaVANips2023}, and exploiting a higher quality dataset.
Similarly to the above, TinyLLaVA \cite{TinyLLaVA2024} presents and analyzes a variety of small-scale VLMs built following the LLaVA 1.5 methodology \cite{ImprovedLLaVACvpr2024}.  
In \cite{ALLaVA2024}, using a comprehensive pipeline for synthetic multimodal data creation, the high-quality ALLaVA dataset is derived and is used to train a series of small-scale VLMs. Similarly, in \cite{ArystanbekovVlmKazakhIeeeAccess2026} a compact VLM is developed specifically for the Kazakh language, by curating and using appropriate vision-language datasets. LLaVA-MORE \cite{LLaVAMore2025} explores different architectural combinations for VLMs, examining their strengths and weaknesses and providing comprehensive insights for their design.
Similarly, in \cite{ImpArxiv2024,BunnyArxiv2024} key design choices for VLMs are investigated and new multimodal training datasets are created.
On the other hand, some works introduce new components to enhance the performance of small-scale VLMs. 
For instance, Sphinx-Tiny \cite{SphinxXPmlr2024} utilizes a mixture-of-visual-experts (MoV) to improve the performance of the model and an enhanced high-quality dataset that allows a single-stage training. In LVIDA \cite{LvidaICCV2025}, a layer-wise multiple injection strategy is introduced, to eliminate unnecessary visual processing and improve the computational efficiency of VLMs.
In MoE-LLaVA \cite{MoELLaVA2026}, the capacity of a small-scale VLM is increased and its performance is improved by stacking MoE encoder layers and using a three-stage process to stabilize training.
Finally, MiniGemini \cite{MiniGemini2026} utilizes a ConvNet to generate high-resolution (HR) images, twin vision encoders for processing LR and HR images, and constructs a new high-quality dataset to train small-scale VLMs.

\textbf{Strategies for condensing models.}
Various model compression techniques have been proposed to reduce memory footprint and accelerate model response during inference.
Quantization is probably the most popular direction for reducing the working size of a VLM. 
Q-VLM \cite{QvlmNips2024} employs entropy as a proxy to optimally partition blocks, with the aim of achieving a good trade-off between discretization errors and rounding function search cost, for post-training quantization.
In \cite{MbqLiCvpr2025}, observing significant sensitivity differences between vision and language modalities, a modality-wise balanced quantization approach is proposed. Another direction (complementary to quantization) for compressing VLMs is pruning. In \cite{YiLinPruneSungIclr2024}, a two-stage coarse-to-fine weight pruning approach for VLMs is proposed, utilizing a global importance score approximated via zero-order gradients.
In \cite{FarinaPruneCVPR2024}, a task-agnostic VLM pruning approach is presented that exploits the magnitude and information flow for 
layer-wise pruning.
Neural Architecture Search (NAS) is yet another direction, searching for models that maximize accuracy while meeting constraints such as latency.
For instance, in \cite{ChittyVenkataIcip2025}, NAS is combined with Low-Rank Adaptation (LoRA), to identify the optimal
LoRA rank for each fully-connected layer in VLMs.

\textbf{Distillation.}
In contrast to directly training smaller network architectures or condensing larger models, KD trains a smaller Student network by transferring knowledge to it from a higher-capacity Teacher model. In \cite{LlavadiArxiv2024} the key factors in the distillation of VLMs are investigated, revealing that most strategies for LLM distillation do not necessarily provide performance gains when applied to VLMs.
The works in \cite{MovekdCvpr2025,LlavaModIclr2025} combine distillation with mixture of experts (MoE) to improve the performance of the base VLMs. 
Specifically, MoVE-KD \cite{MovekdCvpr2025} utilizes a mixture-of-LoRA-experts to efficiently distill the knowledge of multiple vision encoders into a single vision encoder, while LLaVA-MoD \cite{LlavaModIclr2025} integrates a MoE structure into the LLM backbone and introduces preference distillation using appropriate datasets.
Other works aim to improve the alignment and distillation of visual tokens in VLMs.
CompoDistill \cite{CompoDistillIclr2026} explicitly aligns the Student’s visual attention with that of the Teacher, to enhance the Student’s visual perception abilities.
In Allign-KD \cite{AlignKdCvpr2025}, the loss function is designed to additionally distill important cross-modal information from the shallow layers of the Teacher to the Student.
In EM-KD \cite{EmKdAaai2026}, the Manhattan distance and the Hungarian matching algorithm are used to allow spatial alignment of vision logits when the Teacher and the Student utilize vision encoders of different architectures.
LLaVA-KD \cite{LlavaKdICCV2025} presents a three-step distillation framework for small-scale VLMs, along with appropriate losses enforcing logit-level consistency of text and visual tokens.
The above methods (and most KD approaches in the literature) utilize only a single Teacher for performing distillation.
In contrast, the proposed framework extends LLaVA-KD, discussed above, using a cascaded strategy that involves more than one Teachers.

\textbf{Capacity gap in distillation.}
It has been often observed in distillation that the performance of a Student model may not improve, or even degrade, when the so-called capacity gap between the Student and the Teacher is large \cite{BusbridgeDistLawsIcml2025}.
In the past, there have been only a few works that tried to address this limitation of KD.
For instance, in \cite{TeachAssistAaai2020} a new model is firstly created, having a capacity in between that of the Student and the Teacher, that will act as a Teacher Assistant (TA). Then, top-down distillation is performed, i.e., first between the Teacher and the TA, and then between the distilled TA and the Student. In \cite{TaSentimAnalysisDongSsci2023}, the above method is applied to the task of sentiment analysis, using a BERT-based Teacher and a CNN-based Student.
In \cite{TAekdLvColing2024}, knowledge from closed-source multilingual neural machine translation (MNMT) systems is transferred to a Student through a fusion model. In \cite{TanMultiExpertKdIeeeAccess2025}, a distillation method specifically for histopathology data analysis is presented, using more than one experts (i.e., histopathology models) in order to effectively leverage the multi-scale information that is encoded in such data. In \cite{DenseGuideTaIccv2021}, multiple TAs are trained sequentially and used to guide the KD loss of the Student. 
The above methods have been mostly investigated on previous-generation deep networks and datasets (e.g. CNNs, trained on cifar and Imagenet), or developed for very specific application domains (e.g., \cite{TanMultiExpertKdIeeeAccess2025}), without taking into consideration the complexities of transferring them to modern VLMs.  
More importantly, as creating one or more intermediate VLMs (acting as TAs) from scratch is very expensive and typically requires massive training datasets, previous methods that foresee such a step are impractical for VLMs in real-world scenarios.
Moreover, when dealing with medium-scale or larger VLMs as Teachers, the exploitation of multi-Teacher frameworks (such as \cite{DenseGuideTaIccv2021,TAekdLvColing2024}) for KD is impractical: the compute/memory resources required for simultaneously using many VLMs are significant. To this end, we propose a bottom-up approach that improves gradually the performance of the Student, using as TA and Teacher two pre-trained VLMs of different capacity; thus, avoiding the need for training TAs, and avoiding the excessive memory requirements of multi-Teacher strategies.

\section{Proposed Method}
\label{S:proposed_method}

\subsection{Bottom-up cascaded distillation scheme}
\label{SS:CKD}

Let three VLMs $f_s$, $f_a$ and $f_t$, having the role of the Student, the TA and Teacher, respectively, 
\begin{equation}
   f_t, f_a, f_s: \mathbb{R}^{k \times d} \rightarrow \mathbb{R}^{c},
\end{equation}
where $k$ is the maximum length of the input token sequence, $d$ is the dimensionality of the input embeddings, and $c$ is the token vocabulary size. We assume that the TA's capacity (as reflected in its performance in a given task) is in between that of the Teacher and the Student. Also, let $\mathfrak{D}$ be a dataset of multimodal instances for VLM training consisting of two subsets, $\mathfrak{D}_1$ of image-caption pairs, and $\mathfrak{D}_2$ of high-quality instruction-following data. The Teacher and TA models are (pre-)trained using the above dataset for general-purpose visual and language understanding (i.e., aligning and finetuning the already trained vision encoder and LLM).

Then, the following two stages are performed in order to gradually train the Student.

\textbf{Stage 1: TA \textrightarrow Student distillation.}
In this stage the TA and the dataset $\mathfrak{D}$ are used to train the Student via distillation.
The specific KD approach applied at this stage should utilize an appropriate loss, to account for the knowledge transfer of both textual and visual information of the TA to the Student.

\textbf{Stage 2: Teacher \textrightarrow Student distillation.}
Stage 1 is repeated, replacing the TA with the stronger Teacher model, and the original Student with the distilled Student model coming out of Stage 1.

For simplicity of illustration, here we consider only one TA, but the proposed approach could be extended to more than one by introducing additional stages, where in each additional stage a weaker TA is replaced by a stronger one, and the Student is updated with the distilled model coming out of the immediately preceding stage.

In our implementation of this bottom-up cascaded distillation (CKD) scheme, we follow the LLaVA \cite{ImprovedLLaVACvpr2024} architecture for the Student/TA/Teacher VLMs, i.e, each VLM consists of a pretrained vision encoder and LLM backbone, and a vision-language connector. All employed VLMs have vision encoder and vision-language connector of the same architecture, and share the same tokenizer; the TA/Teacher VLMs are pre-trained following the TinyLLaVA approach (as described briefly in Section \ref{ss:TinyLLaVA}). As base KD approach, for performing KD between a pair of models (e.g., TA \textrightarrow Student), LLaVA-KD, a recently-proposed SotA method, is utilized (briefly described in Section \ref{ss:LLaVAKD}). Beyond the above implementation-related choices that are experimentally validated in this work, in principle, the proposed bottom-up CKD scheme could be applied to other families of networks and could be instantiated with a different base KD approach.

\subsection{TinyLLaVA for Teacher and TA training}
\label{ss:TinyLLaVA}

The TinyLLaVA approach is used to align the different components of a VML under the LLaVA framework \cite{TinyLLaVA2024,LiuLLaVANips2023,ImprovedLLaVACvpr2024}. TinyLLaVA
consists of two steps, as follows.

\textbf{Pretraining (PT).}
In this step, the already-trained visual encoder and LLM backbone are kept frozen, and only the vision-language connector is trained using the $\mathfrak{D}_1$ dataset, in order to align the two modalities.
Using the tokenizer, the vision encoder and the textual embedding layer of the VLM, each instance in the dataset can be represented as a pair of embedding and token class sequences,
\begin{equation}
  ( \left[ \mathbf{x}_{1}, \dots, \mathbf{x}_{k} \right], \left[ \mathbf{y}_{1}, \dots, \mathbf{y}_{k} \right]), \label{e:inEmedSequence}
\end{equation}
where  $\mathbf{x}_{i} \in \mathbb{R}^{d}$, $\mathbf{y}_i \in \{0,1\}^{c}$, ($\sum_{j=1}^c y_{i,j} = 1$), are the embedding and one-hot class vector of the $i$th token in the sequence.
At each training step, the forward pass of the VLM is used to generate a corresponding logit sequence for each instance, $ \left[ \mathbf{z}_{1}, \dots, \mathbf{z}_{k} \right]$, $\mathbf{z}_{i} \in \mathbb{R}^{c}$.
A logit vector serves as the input to the final softmax function, providing a corresponding probability vector that serves as the output of the network; this is defined as $\mathbf{p}_{i} = \mbox{softmax}(\mathbf{z}_{i})$.
The above logits with the token class vectors (Eq. \eqref{e:inEmedSequence}) are used in the autoregressive loss $\mathcal{L}_{rg}$ for training the VLM,
\begin{equation}
     \mathcal{L}_{rg} = - \sum_{i=1}^{k-1} \sum_{j=1}^{c} y_{i+1,j} \ln p_{i,j}. \label{e:loss_rg}
\end{equation}

\textbf{Supervised Finetuning (FT).}
At this step, the visual encoder is kept frozen, while the vision-language connector with the LLM backbone are further trained using a high-quality instruction dataset $\mathfrak{D}_2$ and the same autoregressive loss $\mathcal{L}_{rg}$ used in the previous step.

\subsection{LLaVA-KD for Student training}
\label{ss:LLaVAKD}

The LLaVA-KD \cite{LlavaKdICCV2025} approach is used to train the Student at each stage of CKD.
This method consists of three steps, extending TinyLLaVA to the distillation paradigm.

\textbf{Distillation pretraining (DPT).}
As in the PT step of TinyLLaVA, here only the vision-language connector weights are not frozen during training with $\mathfrak{D}_1$.
At each training step the forward pass of the Student and the TA (Stage 1 of CKD) or Teacher (Stage 2 of CKD)  VLM is used to generate logit vectors $\left[ \mathbf{s}_{1}, \dots, \mathbf{s}_{k} \right], \left[ \mathbf{z}_{1}, \dots, \mathbf{z}_{k} \right] \in \mathbb{R}^{c \times k}$, respectively, corresponding to the input embedding sequence.
The following loss \cite{LlavaKdICCV2025} is then used to train the VLM,
\begin{equation}
    \mathcal{L}_{dpt} = \mathcal{L}_{rg} + \tau_1 \mathcal{L}_{td} + \tau_2 \mathcal{L}_{vd} + \tau_3 \mathcal{L}_{vc}, \label{e:loss_dft}
\end{equation}
where $\tau_1, \tau_2, \tau_3$ are the various loss weights, and, $\mathcal{L}_{rg}, \mathcal{L}_{td}, \mathcal{L}_{vd}, \mathcal{L}_{vc}$, are the autoregression, text and visual Kullback–Leibler divergence, and visual cosine divergence loss, respectively.
Specifically, assuming that logit vectors irrelevant to the learning process (e.g., associated with padding or administrative tokens) are masked, and the logit sequences are sorted based on their modality, i.e., the first $m$ logit vectors correspond to visual tokens and the rest to the textual ones, the loss components in Eq. \eqref{e:loss_dft} are defined as,
\begin{eqnarray}
    \mathcal{L}_{rg} = - \sum_{i=1}^{k-1} \sum_{j=1}^{c} y_{i+1,j} \ln q_{i,j},  \\   
    \mathcal{L}_{vd} = \sum_{i=1}^{m} \sum_{j=1}^{c} p_{i,j} \ln \left( \frac{p_{i,j}}{q_{i,j}} \right), \\
    \mathcal{L}_{td} = \sum_{i=m+1}^{k} \sum_{j=1}^{c} p_{i,j} \ln \left( \frac{p_{i,j}}{q_{i,j}} \right), \\
    \mathcal{L}_{vc} = 1- \frac{ \sum_{i=1}^{m} \mathbf{a}_i^T \mathbf{b}_i }{ \sqrt{\sum_{i=1}^{m} \mathbf{a}_i^T \mathbf{a}_i} \sqrt{\sum_{i=1}^{m} \mathbf{b}_i^T \mathbf{b}_i} },
\end{eqnarray}
where, as already defined above $\mathbf{p}_{i} = \mbox{softmax}(\mathbf{z}_{i})$, similarly, $\mathbf{q}_{i} = \mbox{softmax}(\mathbf{s}_{i})$, 
and, $\mathbf{a}_i, \mathbf{b}_i \in \mathbb{R}^{m}$, are the $i$th columns of the autocorrelation matrices of the logit sequences corresponding to visual tokens with respect to the TA (or Teacher) and Student, respectively, i.e., 
\begin{eqnarray}
    \left[ \mathbf{a}_{1}, \dots, \mathbf{a}_{m} \right] &=&  \left[ \mathbf{z}_{1}, \dots, \mathbf{z}_{m} \right]^T \left[ \mathbf{z}_{1}, \dots, \mathbf{z}_{m} \right], \\
    \left[ \mathbf{b}_{1}, \dots, \mathbf{b}_{m} \right] &=&  \left[ \mathbf{s}_{1}, \dots, \mathbf{s}_{m} \right]^T \left[ \mathbf{s}_{1}, \dots, \mathbf{s}_{m} \right].
\end{eqnarray}
The losses $\mathcal{L}_{rg}$, $\mathcal{L}_{td}$ are the common LLM losses used when only the text modality is present, while the losses $\mathcal{L}_{vd}$, $\mathcal{L}_{vc}$ are used to explicitly transfer the rich visual perception abilities from the Teacher to the Student.

\textbf{Supervised finetuning (FT).} This step is the same as in TinyLLaVA (Section \ref{ss:TinyLLaVA}).
The target of this step is to enhance the capability of the Student VLM to understand and follow instructions.

\textbf{Distillation finetuning (DFT).} This step extends the above one by integrating the distillation strategy used in the DPT step. Specifically, the vision-language connector and the LLM backbone are further trained using the overall loss given in Eq. \eqref{e:loss_dft}.
The target of this step is to further enhance the high-level capabilities of the Student VLM using the knowledge of the Teacher.

\subsection{Theoretical analysis}
\label{SS:TheorAnalysis}

Here we provide a theoretical analysis to motivate why progressive reduction of the Teacher-Student gap may improve distillation. 

\textbf{Background.} Based on the statistical learning theory \cite{vapnikStatLearn1998}, the classification error of the Student $f_s$ (or any other classifier) trained directly on a dataset of $n$ instances can be expressed as
\begin{equation}
    R(f_s) - R(f)  \leq O \left( \frac{| \mathcal{F}_s |_C}{ n^{a_s} }  \right) + \epsilon_s, \label{e:s_bound}
\end{equation}
where $\mathcal{F}_s$ is the function class that $f_s$ belongs to, $f \in \mathcal{F}$ is the real target function of interest that we want the Student to learn and $\mathcal{F}$ is the respective function class, $O(\cdot)$ is the estimation error that depends on the statistical learning procedure at hand, $\epsilon_s$ is the approximation error of the Student function class $\mathcal{F}_s$ with respect to $f$, $| \cdot |_C$ is some function class capacity measure, $R(\cdot)$ is the error and $\frac{1}{2} \leq a_s \leq 1$.
The latter is related to the learning rate of the classifier, i.e., for easy problems the exponent is closer to 1, while for difficult ones assumes values closer to $\frac{1}{2}$, translating to fast or slow learning rates, respectively.

In \cite{UnifDistillPaz2016}, a Teacher $f_t \in \mathcal{F}_t$ is introduced to facilitate training through distillation.
Similarly to Eq. \eqref{e:s_bound}, the classification error of the Teacher trained on the same dataset can be similarly expressed as
\begin{equation}
    R(f_t) - R(f)  \leq O \left( \frac{| \mathcal{F}_t |_C}{ n^{a_t} }  \right) + \epsilon_t, \label{e:t_bound}
\end{equation}
where $\epsilon_t$ is the approximation error of the Teacher function, and it is assumed that the Teacher can learn at a faster rate than the Student, i.e. $a_t \geq a_s$.
Then, using the Teacher to train the Student via distillation, an alternative expression for the learning rate of the Student is derived in \cite{UnifDistillPaz2016} 
\begin{equation}
    R(f_s) - R(f) \leq  O \left( \frac{| \mathcal{F}_{s} |_C + | \mathcal{F}_{t} |_C }{ n^{a_{st}}} \right) + \epsilon_{st}  + \epsilon_{t},
     \label{e:st_tighterbound}
\end{equation}
where $\epsilon_{st}$, $a_{st}$, are the approximation error and learning rate of the Student trained via distillation from the Teacher.

\textbf{Cascaded distillation.} In the proposed cascaded distillation approach we assume that a TA is also provided, and the following bottom-up process is followed to train the Student: i) training the TA independently, and then training the Student via distillation using the TA, and ii) training the Teacher independently, and then training the distilled Student via a second distillation using the Teacher.
As in \cite{UnifDistillPaz2016,TeachAssistAaai2020}, in the following we are working with upper bounds and also in an asymptotic region, for distillation we assume pure distillation, and assume the same dataset of $n$ instances is used in all training stages.

Similarly to Eq. \eqref{e:s_bound}, we can express the error of the TA $f_a \in \mathcal{F}_a$, when learning directly from the dataset, as
\begin{equation}
    R(f_a) - R(f)  \leq O \left( \frac{| \mathcal{F}_a |_C}{ n^{a_a} }  \right) + \epsilon_a, \label{e:a_bound} 
\end{equation}
where $\mathcal{F}_a$, $\epsilon_a$, $a_a$ are defined for the TA in analogy to the definitions under Eq. \eqref{e:s_bound}.
Subsequently, the Student learns from the TA via distillation, with error expressed as 
\begin{equation}
    R(f_{s}) - R(f_a)  \leq O \left( \frac{| \mathcal{F}_{s} |_C }{ n^{a_{sa}}}  \right) + \epsilon_{sa}, \label{e:sa_bound}
\end{equation}
where $\epsilon_{sa}$, $a_{sa}$, are the approximation error and learning rate of the Student trained via distillation from the TA.
Similarly to \cite{UnifDistillPaz2016}, by combining Eqs.~\eqref{e:a_bound}, \eqref{e:sa_bound}, an alternative expression can be retrieved for the rate at which the Student learns the truth function,
\begin{eqnarray}
     R(f_{s}) - R(f) &=& R(f_{s}) - R(f_a) + R(f_a) - R(f) \nonumber \\
     &\leq&  O \left( \frac{| \mathcal{F}_{s} |_C }{ n^{a_{sa}}}  \right) + \epsilon_{sa} + O \left( \frac{| \mathcal{F}_{a} |_C }{ n^{a_{a}}}  \right) + \epsilon_{a}, \nonumber \\
     &\leq&  O \left( \frac{| \mathcal{F}_{s} |_C + | \mathcal{F}_{a} |_C }{ n^{a_{sa}}} \right) + \epsilon_{sa}  + \epsilon_{a},
     \label{e:sa_tighterbound}
\end{eqnarray}
where the last inequality holds due to $a_{sa} \leq a_{a}$ (the TA can learn the true distribution at higher rate, compared to the rate that the Student learns the distribution of the TA).

In the next stage the Teacher is trained, and the distilled Student is trained via distillation from the Teacher.
As in \cite{UnifDistillPaz2016}, the learning rate of the Teacher can be expressed as shown in Eq. \eqref{e:t_bound}.
Concerning the distilled Student, we represent her with another function $f_{\bar{s}}$ that can learn at a higher rate $a_{\bar{s}}$, i.e., $a_{s} \leq a_{\bar{s}}$. We base this, following \cite{UnifDistillPaz2016} (see Section 4 of \cite{UnifDistillPaz2016}), on that the Student after distillation has acquired a stronger feature extraction backbone; this can provide features with richer information to the classification head, thus accelerating the learning rate of the model. Consequently, the classification error of the distilled Student $f_{\bar{s}} \in \mathcal{F}_{\bar{s}}$, if it were further trained without distillation, would be expressed as
\begin{equation}
    R(f_{\bar{s}}) - R(f)  \leq O \left( \frac{| \mathcal{F}_{\bar{s}} |_C}{ n^{a_{\bar{s}}} }  \right) + \epsilon_{\bar{s}}, \label{e:ds_bound}
\end{equation}
where $\mathcal{F}_{\bar{s}}$, $\epsilon_{\bar{s}}$, are defined accordingly.
Similarly, for the distillation case, we can write
\begin{equation}
    R(f_{\bar{s}}) - R(f_t)  \leq O \left( \frac{| \mathcal{F}_{\bar{s}} |_C }{ n^{a_{\bar{s}t}}}  \right) + \epsilon_{\bar{s}t}, \label{e:st_bound}
\end{equation}
where $\epsilon_{\bar{s}t}$, $a_{\bar{s}t}$ are the approximation error and learning rate of $f_{\bar{s}}$ trained via distillation from the Teacher.
Combining Eqs.~\eqref{e:t_bound}, \eqref{e:st_bound}, and noting that $a_{\bar{s}t} \leq a_{t}$, an alternative expression for the learning rate of the Student is retrieved
\begin{eqnarray}
    R(f_{\bar{s}}) - R(f) &=& R(f_{\bar{s}}) - R(f_t) + R(f_t) - R(f) \nonumber \\
    &\leq&  O \left( \frac{| \mathcal{F}_{\bar{s}} |_C }{ n^{a_{\bar{s}t}}}  \right) + \epsilon_{\bar{s}t} + O \left( \frac{| \mathcal{F}_{t} |_C }{ n^{a_{t}}}  \right) + \epsilon_{t} \nonumber \\
     &\leq&  O \left( \frac{| \mathcal{F}_{\bar{s}} |_C + | \mathcal{F}_{t} |_C }{ n^{a_{\bar{s}t}}} \right) + \epsilon_{\bar{s}t}  + \epsilon_{t},
     \label{e:dst_tighterbound}
\end{eqnarray}
Based on Eqs.~\eqref{e:st_tighterbound}, \eqref{e:dst_tighterbound}, in order for CKD to outperform the baseline KD (directly from the Teacher), we need to show that the following inequality is satisfied:
\begin{eqnarray}
    O \left( \frac{| \mathcal{F}_{\bar{s}} |_C + | \mathcal{F}_{t} |_C }{ n^{a_{\bar{s}t}}} \right) + \epsilon_{\bar{s}t} &\leq& O \left( \frac{| \mathcal{F}_{s} |_C + | \mathcal{F}_{t} |_C }{ n^{a_{st}}} \right) + \epsilon_{st} \nonumber \\ \label{e:dst_vs_st} 
\end{eqnarray}
Considering that the gap between the distilled Student and the Teacher is smaller than the gap between the Teacher and the Student prior to performing distillation with the TA  (as the distilled Student can leverage a richer feature representation, in comparison to the Student \cite{UnifDistillPaz2016}), we can write $a_{st} \leq a_{\bar{s}t}$, establishing the validity of Eq. \eqref{e:dst_vs_st} in the asymptotic regime.

\section{Experiments}
\label{S:Exp}

\subsection{Experimental Setup}
\label{SS:ExpSetup}

\textbf{Implementation Details.}
Training of the Student in each stage of LLaVA-CKD is performed following the settings of our baseline KD method, LLaVA-KD.
More specifically, standard AdamW optimizer is used with cosine decay scheduler and warm-up ratio of 3$\times$10\textsuperscript{-2}, and the training is performed for one epoch at each stage.
For the DPT step (Section \ref{ss:LLaVAKD}) the initial learning rate and batch size are set to 10\textsuperscript{-3} and 256, while for the SFT and DFT steps (Section \ref{ss:LLaVAKD}) to
2$\times$10\textsuperscript{-5} and 128, respectively.
Concerning the weights in the LLaVA-KD loss (Eq. \eqref{e:loss_dft}), for simplicity all weights are set to one.
The method is implemented\footnote{The source code implementation of LLaVA-CKD and the distilled models evaluated in the Experiments Section are available at \url{https://github.com/IDT-ITI/LLaVA-CKD}} using Python and appropriate packages such as PyTorch and DeepSpeed.
For the execution of the experiments, a workstation with two Nvidia RTX PRO 6000 GPUs and 256 GB RAM was utilized.

\textbf{Base Models and Training Datasets.}
In all experiments with the proposed method and our baselines, the VLM consisted of a pretrained SigLIP model (\text{SigLIPB/14@384px}) \cite{ZhaiSigClipIccv2023} and an LLM of the Qwen2.5 series \cite{Qwen25TechReprt2024} as Visual Encoder and LLM backbone, respectively.
As vision-language connector we use a two-layer MLP with GELU activation function (\text{MLP2x\_GELU}), which is randomly initialized prior to the first training.
The LLaVA1.5 dataset \cite{ImprovedLLaVACvpr2024} is used for training the various methods, consisting of two subsets: i) the subset LLaVA1.5-558k with image-caption pairs (denoted $\mathfrak{D}_1$ in the PT and DPT step in Sections \ref{ss:TinyLLaVA}, \ref{ss:LLaVAKD}), and, ii) the high-quality instruction dataset LLaVA-mix-665k (denoted $\mathfrak{D}_2$ in the FT and DFT step in Sections \ref{ss:TinyLLaVA}, \ref{ss:LLaVAKD}).

\textbf{Baselines.}
The proposed LLaVA-CKD is compared with our main baselines, i.e., TinyLLaVA \cite{TinyLLaVA2024}, which is used to build our VLM models, and LLaVA-KD \cite{LlavaKdICCV2025}, which is utilized as our base distillation method. For experimentation, the models are built using the official codes and the settings provided in the corresponding papers, and as in \cite{TinyLLaVA2024,LlavaKdICCV2025} they are trained for one epoch at each stage.

\textbf{Comparison with SotA.}
We compare LLaVA-CKD with SotA methods for creating compact VLMs (both KD-based and others) that reported using a comparable-size LLM backbone (either in the 0.5B or the 1.5B parameters class, matching the Student LLM backbones we use in the LLaVA-CKD experiments) and a comparable number of training samples:
LLAVADI \cite{LlavadiArxiv2024},
LLaVA-MoD \cite{LlavaModIclr2025},
MoVE-KD \cite{MovekdCvpr2025},
Allign-KD \cite{AlignKdCvpr2025},
CompoDistill \cite{CompoDistillIclr2026},
MiniGemini \cite{MiniGemini2026},
Sphinx-Tiny \cite{SphinxXPmlr2024},
Bunny \cite{BunnyArxiv2024},
Imp \cite{ImpArxiv2024},
ALLaVA \cite{ALLaVA2024},
EM-KD \cite{EmKdAaai2026},
LLaVA-MORE \cite{LLaVAMore2025},
MoE-LLaVA \cite{MoELLaVA2026},
LVIDA \cite{LvidaICCV2025}.

\begin{table*}[t]
  \centering
  \caption{Comparison between our baselines (TinyLLaVA, LLaVA-KD) and the proposed method (LLaVA-CKD) with Student LLM backbone Qwen2.5-0.5B. The TinyLLaVA models used as Teacher and/or TA in all distillation experiments reported in this table are presented in the first (Qwen2.5-3B) and second (Qwen2.5-1.5B) block of the table. The results for TinyLLaVA and LLaVA-KD have been reproduced using the official code. In the shaded rows, we also present the corresponding results as originally reported in \cite{LlavaKdICCV2025}. Best and second-best results for the Student model (Qwen2.5-0.5B; third block of the table), for the experiments executed under the exact same conditions (i.e., excluding the shaded rows) are shown in bold and underline, respectively. Improvements over the TinyLLaVA baseline, in terms of average scores, are also indicated in green.}
\begin{tabular}{lccccccccl}\toprule
\textbf{Method} & \textbf{LLM size(s)} & VQA\textsuperscript{v2} & GQA & VQA\textsuperscript{T} & SQA\textsuperscript{I} & MME\textsuperscript{P} & POPE & MMMU & \textbf{Average} \\
\midrule
\rowcolor{gray!10} TinyLLaVA & 3B & 80.4 & 63.2 & 61.5 & 76 & 73.9 & 85.9 & 40.3 & 68.7 \\ 
TinyLLaVA \textcolor{gray}{\tiny ARXIV'24} (Teacher) & \dittoclosing & 79.4 & 62.5 & 58.3 & 74.1 & 72.3 & 86.6 & 40.2 & 67.6 \\
\midrule
\rowcolor{gray!10} TinyLLaVA & 1.5B & 78.8 & 62 & 57.4 & 72 & 72.5 & 85.5 & 37 & 66.5 \\
TinyLLaVA \textcolor{gray}{\tiny ARXIV'24} (Teacher/TA) & \dittoclosing & 76.9 & 57.7 & 56.4 & 71.7 & 69.2 & 85.7 & 37.4 &  65.0 \\
\midrule
\rowcolor{gray!10} TinyLLaVA & 0.5B & 74.8 & 58.3 & 49.2 & 59.1 & 61.5 & 86.1 & 33.6 & 60.4 \\
TinyLLaVA \textcolor{gray}{\tiny ARXIV'24} (Student) & \dittoclosing & 74.3 & 55 & \ul{49.2} & 60.9 & 60.7 & 85.5 & \textbf{33.1} &  59.8 \textcolor{green}{(+0.0)}\\
LLaVA-KD \textcolor{gray}{\tiny ICCV'25} & 1.5B \textrightarrow 0.5B & 76.2 & 58.6  & 48.1 & \ul{61.1} & 63.1 & 85.8 & 30.8 & 60.5 \textcolor{green}{(+0.7)} \\
\rowcolor{gray!10} LLaVA-KD  & 3B \textrightarrow 0.5B & 77.7 & 59.8 & 52 & 60.6 & 64.7 & 86.4 & 28.3 & 61.2 \\ 
LLaVA-KD \textcolor{gray}{\tiny ICCV'25} & \dittoclosing & \ul{76.9} & \ul{60.2} & 48.5 & 59.9 & \ul{64.6} & \textbf{86.6} & 29.9 & \ul{60.9} \textcolor{green}{(+1.1)} \\
LLaVA-CKD (proposed) & 1.5B \textrightarrow 0.5B; 3B \textrightarrow 0.5B & \textbf{77.9} & \textbf{60.5} & \textbf{51} & \textbf{61.6} & \textbf{66.2} & \ul{86.4} & \ul{30.9} & \textbf{62.1} \textcolor{green}{(+2.3)} \\
\bottomrule
\end{tabular}
  \label{tab:baselines_05b}%
\end{table*}%

\begin{table*}[t]
  \centering
  \caption{Comparison between the proposed method (LLaVA-CKD) and SotA small-scale VLMs with an LLM backbone equal to 0.5B, or at least smaller than 1.5B. For each method, the number of used training samples is shown in column \#TS. Methods belonging to the KD category are denoted with green highlight. Best and second-best results are shown in bold and underline, respectively.}
\begin{tabular}{lcccccccccc}\toprule
\textbf{Method} & \textbf{LLM} & \textbf{\#TS} & VQA\textsuperscript{v2} & GQA & VQA\textsuperscript{T} & SQA\textsuperscript{I} & MME\textsuperscript{P} & POPE & MMMU \\
\midrule
\rowcolor{green!10} LLAVADI \textcolor{gray}{\tiny ARXIV'24} & MLLaMA-1.4B & 1.2M & - & 55.4 & 45.3 & 56 & 58.9 & 84.7 & -  \\
Sphinx-Tiny \textcolor{gray}{\tiny PMLR'24} & TinyLlama-1.1B & 15M & 74.7 & 58 & \textbf{57.8} & 49.2 & 63.1 & 82.2 & -  \\
LVIDA \textcolor{gray}{\tiny ICCV'25} & Llama-3.2-1B & 1.2M & \ul{76.2} & \ul{59.6} & 50.6 & - & 60 & \textbf{86.6} & 29.9 \\
\rowcolor{green!10} MoVE-KD-v1.1 \textcolor{gray}{\tiny CVPR'25} & MobLLaMA-1.4B & 1.2M & 73.8 & 57.7 & 44.3 & 57.3 & 59.4 & 86.1 & - \\
\rowcolor{green!10} LLaVA-MoD \textcolor{gray}{\tiny ICLR'25} & Qwen1.5-0.5B & 5M & - & 56.2 & \ul{53.9} & \textbf{62.8} & \ul{65.3} & - & - \\
\rowcolor{green!10} EM-KD \textcolor{gray}{\tiny AAAI'26} & Qwen2.5-0.5B & 1.3M & 57.8 & & 49.2 & 53 & 61.3 & - & \textbf{31} \\
\rowcolor{green!10} LLaVA-CKD  \textcolor{gray}{(proposed)} & Qwen2.5-0.5B & 1.2M &  \textbf{77.9} & \textbf{60.5} & 51 & \ul{61.6} & \textbf{66.2} & \ul{86.4} & \ul{30.9} \\
\bottomrule
\end{tabular}
  \label{tab:SOTA_05b}%
\end{table*}%

\textbf{Evaluation Protocol.}
The experiments are conducted using seven publicly available benchmarks for VLM evaluation: i) VQA v2 (VQA\textsuperscript{v2}) \cite{GoyalVqaV2CVPR2017} for scene understanding and compositionality, ii) GQA \cite{HudsonGqaCvpr2019} for general visual understanding and relational reasoning, iii) TextVQA (VQA\textsuperscript{T}) \cite{SinghTextVqaCvpr2019} for fine-grained visual recognition and understanding of text within images, iv)  
ScienceQA-IMG (SQA\textsuperscript{I}) \cite{LuScienceQANips2022} to measure understanding of scientific knowledge, v) MME Perception (MME\textsuperscript{P}) \cite{MmeArxiv2023} for comprehensive evaluation of multimodal understanding and reasoning capabilities, vi) POPE \cite{LiPopeEMNLP2023} for hallucination, and, vii) MMMU \cite{MmmuYueCVPR2024} for multi-disciplinary tasks of college-level education. In accordance with standard practice for these datasets, accuracy is used as the evaluation measure for all except MME\textsuperscript{P} and POPE; for the former the perception score (which is a measure of accuracy along a set of subtasks) is computed and normalized by dividing it with 2000 (the maximum possible score), while for the latter, the average F1-score over its three sub-tasks (``Random'', ``Popular'' and ``Adversarial'') is reported. Finally, we should note that the evaluation in VQA\textsuperscript{v2} is done by submitting the obtained results to the server provided online by the respective team. To facilitate comparison of LLaVA-CKD with our baselines, for each method we also report the average of the scores achieved along all benchmarks (column ``Average'' in Tables \ref{tab:baselines_05b}, \ref{tab:baselines_15b}, \ref{tab:ablQwen2Ta} and \ref{tab:ablTdVsBu}), similarly to other works in this domain; combined, these benchmarks total more than 130K testing question instances. Following the typical TinyLLaVA evaluation setup, as in e.g. \cite{TinyLLaVA2024,LlavaKdICCV2025}, we set the VLMs' LLM-backbone temperature to zero in all experiments, triggering greedy decoding (thus, removing randomness and making the models’ output repeatable). Additionally, due to the high computational cost of each training / distillation experiment, we perform each experiment once and report the corresponding result (i.e., we cannot report confidence intervals); this is in line with what all the compared baselines and SotA methods have reported in \cite{TinyLLaVA2024,LlavaKdICCV2025,LlavadiArxiv2024,LlavaModIclr2025,MovekdCvpr2025,AlignKdCvpr2025,CompoDistillIclr2026,MiniGemini2026,SphinxXPmlr2024,BunnyArxiv2024,ImpArxiv2024,ALLaVA2024,EmKdAaai2026,LLaVAMore2025,MoELLaVA2026,LvidaICCV2025}.

\subsection{Main Results and Comparisons}
\label{SS:MainRes}

\textbf{Student LLM backbone: Qwen2.5-0.5B.}
In this set of experiments, we produce a distilled model based on Qwen2.5-0.5B. The TinyLLaVA models with Qwen2.5-1.5B and Qwen2.5-3B LLM backbones are used as TA and Teacher, respectively, in LLaVA-CKD (and simply as Teachers in LLaVA-KD).
The evaluation results of the proposed LLaVA-CKD against the baselines, TinyLLaVA and LLaVA-KD, are depicted in the bottom part of Table \ref{tab:baselines_05b}. We should note that, by definition (Section \ref{SS:CKD}), LLaVA-KD 1.5B \textrightarrow 0.5B constitutes Stage 1 of the LLaVA-CKD experiment reported in this table (thus, the results reported for LLaVA-KD 1.5B \textrightarrow 0.5B also expose the contribution of the TA-only 1st Stage of LLaVA-CKD in its final results). For completeness, we also provide the evaluation results of the Teacher and TA in the first and second blocks of the same table. The results corresponding to the baselines have been obtained using the official codes provided by the authors of TinyLLaVA and LLaVA-KD.
Additionally, in the shaded rows we show the corresponding results reported in \cite{LlavaKdICCV2025} (except for LLaVA-KD 1.5B \textrightarrow 0.5B, as this experiment was not reported in \cite{LlavaKdICCV2025}).

From the results in Table \ref{tab:baselines_05b}, we observe that the proposed LLaVA-CKD outperforms both baselines. Specifically, a notable average performance gain of +1.2 and +1.6 percentage points is observed against LLaVA-KD with Teacher-Student 3B \textrightarrow 0.5B and 1.5B \textrightarrow 0.5B, respectively, and an overall improvement of +2.3 percentage points over TinyLLaVA 0.5B (i.e., the original model without applying any form of distillation).
Looking at the individual datasets, LLaVA-CKD provides the best results in five of the seven benchmarks. In the remaining two datasets, it provides the second-best result; specifically, in POPE, it is slightly outperformed by LLaVA-KD 3B \textrightarrow 0.5B, while in MMMU LLaVA-CKD is second only to the TinyLLaVA baseline.

\begin{table*}[t]
  \centering
  \caption{Comparison between our baselines (TinyLLaVA, LLaVA-KD) and the proposed method (LLaVA-CKD) with Student LLM backbone Qwen2.5-1.5B. The TinyLLaVA models used as Teacher and/or TA in all distillation experiments reported in this table are presented in the first (Qwen2.5-7B) and second (Qwen2.5-3B) block of the table. The results for TinyLLaVA and LLaVA-KD have been reproduced using the official code. In the shaded rows, we also present the corresponding results as originally reported in \cite{LlavaKdICCV2025}. Best and second-best results for the Student model (Qwen2.5-1.5B; third block of the table), for the experiments executed under the exact same conditions (i.e., excluding the shaded rows) are shown in bold and underline, respectively. Improvements over the TinyLLaVA baseline, in terms of average scores, are also indicated in green.}
\begin{tabular}{lccccccccl}\toprule
\textbf{Method} & \textbf{LLM size(s)} & VQA\textsuperscript{v2} & GQA & VQA\textsuperscript{T} & SQA\textsuperscript{I} & MME\textsuperscript{P} & POPE & MMMU & \textbf{Average} \\
\midrule
\rowcolor{gray!10} TinyLLaVA & 7B & 81.8 & 64.3 & 64.6 & 77.3 & 78.5 & 86.6 & 46.4 & 71.4 \\ 
TinyLLaVA \textcolor{gray}{\tiny ARXIV'24} (Teacher) & \dittoclosing &  80.9 & 63.4 &  63.6 & 78.4 & 79.8 & 87.5 & 43.9 & 71.1 \\
\midrule
\rowcolor{gray!10} TinyLLaVA & 3B & 80.4 & 63.2 & 61.5 & 76 & 73.9 & 85.9 & 40.3 & 68.7 \\ 
TinyLLaVA \textcolor{gray}{\tiny ARXIV'24} (Teacher/TA) & \dittoclosing & 79.4 & 62.5 & 58.3 & 74.1 & 72.3 & 86.6 & 40.2 & 67.6 \\
\midrule
\rowcolor{gray!10} TinyLLaVA & 1.5B & 78.8 & 62 & 57.4 & 72 & 72.5 & 85.5 & 37 & 66.5 \\
TinyLLaVA \textcolor{gray}{\tiny ARXIV'24} (Student) & \dittoclosing & 76.9 & 57.7 & 56.4 & 71.7 & 69.2 & 85.7 & 35.4 &  65.0 \textcolor{green}{(+0.0)} \\
\rowcolor{gray!10} LLaVA-KD & 3B \textrightarrow 1.5B & 80.3 & 62.5 & 59.7 & 71.6 & 70 & 86.7 & 35.8 & 66.7 \\
LLaVA-KD \textcolor{gray}{\tiny ICCV'25} & \dittoclosing & 79.4 & \textbf{61.6} & 58.7 & 71.9 & 69.9 & \textbf{86.7} & \textbf{37.3} & \ul{66.5} \textcolor{green}{(+1.5)} \\
LLaVA-KD \textcolor{gray}{\tiny ICCV'25} & 7B \textrightarrow 1.5B & \ul{79.8} & 61.3 & \ul{59} & \ul{72.3} & \ul{70.1} & 85.8 & 35.9 & 66.3 \textcolor{green}{(+1.3)} \\
LLaVA-CKD (proposed) & 3B \textrightarrow 1.5B; 7B \textrightarrow 1.5B & \textbf{80} & \ul{61.4} & \textbf{60} & \textbf{72.6} & \textbf{73.7} & \ul{85.9} & \ul{37} & \textbf{67.2} \textcolor{green}{(+2.2)} \\
\bottomrule
\end{tabular}
  \label{tab:baselines_15b}%
\end{table*}%

\begin{table*}[t]
  \centering
  \caption{Comparison between the proposed method (LLaVA-CKD) and SotA small-scale VLMs with an LLM backbone with 1.5B to 2B parameters. For each method, the number of used training samples is shown in column \#TS. Methods belonging to the KD category are denoted with green highlight. Best and second-best results are shown in bold and underline, respectively.}
\begin{tabular}{lcccccccccc}\toprule
\textbf{Method} & \textbf{LLM} & \textbf{\#TS} & VQA\textsuperscript{v2} & GQA & VQA\textsuperscript{T} & SQA\textsuperscript{I} & MME\textsuperscript{P} & POPE & MMMU \\
\midrule
ALLaVA \textcolor{gray}{\tiny ARXIV'24} & StableLM-2-1.6B & 1.2M & - & 49.8 & 51.7 & 64.7 & 65.6 & - & 33.3 \\
Bunny \textcolor{gray}{\tiny ARXIV'24} & Qwen1.5-1.8B & 2.6M & 76.6 & 59.6 & 53.2 & 64.6 & 65 & 85.8 & - \\
Imp \textcolor{gray}{\tiny ARXIV'24} & Qwen1.5-1.8B & 1.5M & \ul{79.2} & 61.9 &  54.5 & 66.1 & 65.2 & \ul{86.7} & - \\
\rowcolor{green!10} LLaVA-MoD \textcolor{gray}{\tiny ICLR'25} & Qwen1.5-1.8B & 5M & - & 58.7 & \ul{58.5} & 68 & 66.7 & - & - \\
\rowcolor{green!10} Allign-KD \textcolor{gray}{\tiny CVPR'25} & MobVLM\textsuperscript{v2}-1.7B & 3.6M & - & 60.1 & 53.1 & 67.7 & 65.2 & \textbf{87} & - \\
MoE-LLaVA \textcolor{gray}{\tiny TMM'26} & Qwen1.5-1.8B & 2.1M & 76.2 & 61.5 & 48 & 63.1 & 64.6 & \textbf{87} & - \\
LLaVA-MORE \textcolor{gray}{\tiny ICCVW'26} & Gemma-2-2B & 1.2M & - & \textbf{62.4} & 54.4 & \ul{71.1} & - & 86 & \ul{33.4}  \\
MiniGemini \textcolor{gray}{\tiny TPAMI'26} & Gemma-2B & 2.2M &  & 60.7 & 56.2 & 63.1 & \ul{67} & 85.6 & 31.7 \\
\rowcolor{green!10} CompoDistill \textcolor{gray}{\tiny ICLR'26} & Qwen1.5-1.8B & 1.2M & 78.8 & \ul{62.2} & 56.4 & 70.1 & - & - & - \\
\rowcolor{green!10} LLaVA-CKD \textcolor{gray}{(proposed)} & Qwen2.5-1.5B & 1.2M & \textbf{80} & 61.4 & \textbf{60} & \textbf{72.6} & \textbf{73.7} & 85.9 & \textbf{37} \\
\bottomrule
\end{tabular}
  \label{tab:SOTA_15b}%
\end{table*}%

Overall, we observe that distillation has a beneficial effect to the performance of the model in all datasets except for MMMU, where a performance drop, compared to the TinyLLaVA baseline model, is observed when any kind of distillation is applied. The same conclusion can also be drawn from the results reported in \cite{LlavaKdICCV2025}. This may be attributed to the fact that the training dataset contains few instances relevant to the college-level tasks found in MMMU, which in addition to the relatively small capacity of the 0.5B model, leads to some catastrophic forgetting in the resulting distilled models for this type of knowledge; however, investigating the possible misalignment between the MMMU tasks and the standard training data of \cite{ImprovedLLaVACvpr2024} is beyond the scope of this work.

The comparison of LLaVA-CKD against the SotA methods of the literature for generating similarly small-scale VLMs (with an LLM backbone equal to 0.5B, or at least smaller than 1.5B), is reported in Table \ref{tab:SOTA_05b}. We should note that this comparison involves compact VLMs generated under heterogeneous experimental conditions, as indicated in the ``LLM'' and ``\#TS'' columns of this Table. We observe that the proposed approach provides very competitive performance. Specifically, LLaVA-CKD outperforms all other methods in three out of the seven datasets, and is second-best in another three datasets; no single compared method achieves first place in more than one of the seven datasets. In one last dataset, VQA\textsuperscript{T}, LLaVA-CKD achieves the third-best result. However, we observe from Table \ref{tab:SOTA_05b} that the two methods that outperform it in this dataset, Sphinx-Tiny and LLaVA-MoD, utilize much larger training datasets of 15M and 5M instances, respectively. The wealth of additional training data, which are clearly well-aligned with the VQA\textsuperscript{T} dataset, explain the exceptional performance of these methods specifically in this dataset. More generally, we should note that, as clearly indicated in this Table, inevitably not all compared SotA methods use the same LLM backbone and training data; for this reason, we must acknowledge that the observed performance differences between LLaVA-CKD and the compared methods may be affected, either positively or negatively, by these implementation differences. E.g., the LLaVA-MoD method may have a performance disadvantage due to using a previous-generation Qwen1.5 LLM, but at the same time, using a much larger training dataset (5M training samples instead of 1.2M) compared to LLaVA-CKD gives it a clear performance boost. Similarly, several compared methods use a larger than 0.5B LLM backbone (within the 0.5B to 1.5B range), and their performance may be boosted to some extent by their increased LLM size compared to LLaVA-CKD. Even with these considerations in mind, Table \ref{tab:SOTA_05b} clearly documents the very competitive performance of LLaVA-CKD.

\textbf{Student LLM backbone: Qwen2.5-1.5B.}
In this second set of experiments, we produce a distilled model based on the larger (but still small-scale) Qwen2.5-1.5B LLM. The TinyLLaVA models with Qwen2.5-3B and Qwen2.5-7B LLM backbones are used as TA and Teacher, respectively, in LLaVA-CKD (and as Teachers in LLaVA-KD). The evaluation results for the Teacher, TA, and the proposed method together with the baselines used for comparison are shown in the first, second, and third block of Table \ref{tab:baselines_15b}, respectively.
As in the previous set of experiments, the results of our baselines as reported in \cite{LlavaKdICCV2025} are also shown, in the shaded rows.

From the results in Table \ref{tab:baselines_15b}, we observe that LLaVA-CKD again outperforms both baselines, with a notable average performance gain of +0.7 percentage points over the second-best method (LLaVA-KD 3B \textrightarrow 1.5B) and a leap of +2.2 percentage points over the TinyLLaVA model. With respect to the individual datasets, LLaVA-CKD provides the best results in four of the seven benchmarks and the second-best result in the remaining three. It is also interesting to note that the performance of the proposed method (Average: 67.2\%) almost reaches the performance of the TinyLLaVA model in the larger 3B scale (67.6\%), exhibiting the potential of LLaVA-CKD to close the gap between different model scales.
Another interesting observation is that LLaVA-KD (our KD baseline) in the 1.5B scale provides slightly better results when using the smaller Teacher (3B) rather than the larger one (7B); we attribute this to the large capacity gap between the stronger Teacher and the Student, in line with the findings of \cite{BusbridgeDistLawsIcml2025}.

The comparison of LLaVA-CKD against the SotA methods of the literature for generating VLMs in the same model scale (i.e., with an LLM backbone having between 1.5B and 2B parameters), is reported in Table \ref{tab:SOTA_15b}. Similarly to Table \ref{tab:SOTA_05b}, this comparison involves compact VLMs generated under heterogeneous experimental conditions; inevitably, again not all compared SotA methods use the same LLM backbone and training data. For this reason, the observed performance differences between LLaVA-CKD and the compared methods may be affected, either positively or negatively, by differences in LLM backbone generation, LLM size (all compared methods use somewhat larger LLMs than the 1.5B model used in LLaVA-CKD), and number of employed training samples (most compared methods use more than the 1.2M training samples used by LLaVA-CKD). Even so, we observe from Table \ref{tab:SOTA_15b} that LLaVA-CKD provides very competitive results, outperforming all other methods in five out of the seven datasets. In several of these datasets LLaVA-CKD outperforms the second-best approach by a large margin, i.e., in MME\textsuperscript{P}, MMMU and VQA\textsuperscript{T}. 

\begin{table*}[t]
  \centering
  \caption{Ablation of using a TA of a different Qwen generation (Qwen2 instead of Qwen2.5).}
\begin{tabular}{lccccccccc}\toprule
\textbf{Method} & \textbf{LLM size(s)} & VQA\textsuperscript{v2} & GQA & VQA\textsuperscript{T} & SQA\textsuperscript{I} & MME\textsuperscript{P} & POPE & MMMU & \textbf{Average} \\
\midrule
TinyLLaVA (TA) & 1.5B (Qwen2) & 77.7 & 59.6 & 56.1 & 70.9 & 69.2 & 86.3 & 35.8 & 65.1 \\
\midrule
LLaVA-KD & 1.5B (Qwen2) \textrightarrow 0.5B & 76.6 & 58.1 & 50.6 & 59.6 & 62 & 85.6 & 29.8 & 60.3 \textcolor{green}{(+0.5)} \\
LLaVA-CKD (proposed) & 1.5B (Qwen2) \textrightarrow 0.5B; 3B \textrightarrow 0.5B & \textbf{76.9} & \textbf{58.7} & \textbf{51.4} & \textbf{61} & \textbf{65.2} & \textbf{86.7} & \textbf{32.8} & \textbf{61.8} \textcolor{green}{(+2.0)} \\
\bottomrule
\end{tabular}
  \label{tab:ablQwen2Ta}%
\end{table*}%

\begin{table*}[t]
  \centering
  \caption{Ablation of different cascaded distillation strategies: Top-down cascaded distillation vs repeated same-teacher LLaVA-KD (LLaVA-KD $\times 2$) vs bottom-up cascaded distillation (i.e., the proposed LLaVA-CKD).}
\begin{tabular}{lccccccccc}\toprule
\textbf{Distillation strategy} & \textbf{LLM size(s)} & VQA\textsuperscript{v2} & GQA & VQA\textsuperscript{T} & SQA\textsuperscript{I} & MME\textsuperscript{P} & POPE & MMMU & \textbf{Average} \\
\midrule
Top-down & 3B \textrightarrow  1.5B \textrightarrow  0.5B  & 76.2 & 58.2 & \textbf{51} & \ul{61.4} & 61.9 & \textbf{86.5} & \textbf{31.9} & 61.0 \textcolor{green}{(+1.2)}\\
LLaVA-KD $\times 2$ & 3B \textrightarrow  0.5B; 3B \textrightarrow  0.5B  & \textbf{78.3} & \textbf{60.7} & 50.2 & 61 & \ul{65} & 85.8 & \ul{31.7} & \ul{61.8} \textcolor{green}{(+2.0)}\\
LLaVA-CKD (proposed, bottom-up) & 1.5B \textrightarrow  0.5B; 3B \textrightarrow  0.5B  & \ul{77.9} & \ul{60.5} &  \textbf{51} & \textbf{61.6} & \textbf{66} & \ul{86.4} & 30.9 & \textbf{62.1} \textcolor{green}{(+2.3)} \\
\bottomrule
\end{tabular}
  \label{tab:ablTdVsBu}%
\end{table*}%

\subsection{Ablation Studies}
\label{SS:AblStudy}

\textbf{Using as TA a model from another LLM family.}
In this ablation we repeat the LLaVA-CKD experiment 3B \textrightarrow 1.5B \textrightarrow 0.5B (Section \ref{SS:MainRes}), however, this time using a Qwen2-based TA \cite{Qwen2TechReprt2024} instead of homogeneously using Qwen2.5 LLMs in all roles within LLaVA-CKD. Specifically, the Qwen2.5-1.5B LLM backbone of the TA is replaced with the Qwen2-1.5B one. The Teacher and Student models remain the same (Qwen2.5-based) as in the main experiments. With this ablation we want to assess the robustness of LLaVA-CKD to such a moderate change in the TA: a TA trained on different data, but still sharing the same basic structure and characteristics (e.g., the same tokenizer) with the Teacher and Student models. The results are shown in Table \ref{tab:ablQwen2Ta}; 
again, improvements over the TinyLLaVA baseline for the same 0.5B model (whose results have been reported in Table \ref{tab:baselines_05b}) in terms of average scores are indicated in green in the last column of Table \ref{tab:ablQwen2Ta}.
From these results we observe that the application of distillation is beneficial even when such a different TA is used. The baseline LLaVA-KD achieves a moderate improvement of +0.5 percentage points over the comparable TinyLLaVA model, while LLaVA-CKD outperforms both baselines by a large margin, i.e., by +2 and +1.5 percentage points over TinyLLaVA and LLaVA-KD, respectively. However, it slightly under-performs in comparison to our main LLaVA-CKD setting where the Qwen2.5-based TA is used (average score 61.8\% for this ablation, vs. 62.1\% for the main LLaVA-CKD setting reported in Table \ref{tab:baselines_05b}).

\textbf{Top-down vs bottom-up distillation.}
We compare the proposed approach that follows a bottom-up distillation strategy (i.e., 1.5B \textrightarrow 0.5B; 3B \textrightarrow 0.5B), against a similar implementation of a top-down distillation strategy, i.e., a strategy analogous to the approaches proposed in \cite{TeachAssistAaai2020,DenseGuideTaIccv2021,TaSentimAnalysisDongSsci2023}.
That is, performing KD for one epoch to first train the TA using the Teacher, and then training the Student using the distilled TA for a second epoch, denoted as: Teacher \textrightarrow TA \textrightarrow Student.
In contrary to the bottom-up strategy, in this case the TA also undergoes KD, and the Student is distilled only once (by the distilled TA). The results are shown in Table \ref{tab:ablTdVsBu} (1st row); improvements over the TinyLLaVA baseline for the same 0.5B model (whose results have been reported in Table \ref{tab:baselines_05b}) in terms of average scores are indicated in green in the last column of Table \ref{tab:ablTdVsBu}. We see that the top-down distillation strategy exhibits a good +1.2 percentage points improvement in average performance over the TinyLLaVA baseline, also marginally outperforming the single-stage LLaVA-KD (results reported in Table \ref{tab:baselines_05b}). However, the proposed LLaVA-CKD bottom-up strategy clearly outperforms the top-down one, with an additional +1.1 percentage points gain in average performance (62.1\% vs. 61\%; last column of Table \ref{tab:ablTdVsBu}).  
Additionally, we should note that the top-down strategy has significantly higher computational cost due to its first stage, where the TA (that uses the Qwen2.5-1.5B LLM backbone) serves as a Student (e.g., see \cite{ZeroIeeePress2020} and, for illustration, the middle diagram of Fig. \ref{fig:LlavaCkdConcept}). In contrast, in both stages of LLaVA-CKD the lightweight 0.5B LLM backbone is used as Student, resulting in a relatively lower computational cost for training just this smaller model.

\textbf{Computational cost concerns and equal wall-time comparison.}
The computational cost of the proposed CKD approach is generally proportional to the number of distillation stages it introduces. Compared to the base LLaVA-KD, distillation using LLaVA-CKD with one TA requires approximately twice the compute time (12-15 days on a dual RTX PRO 6000 workstation). However, the benefit is a stronger compact VLM, e.g. from Table \ref{tab:baselines_15b}, a 1.5B model that almost reaches the performance of the base TinyLLaVA 3B model (67.2\% vs. 67.6\%). Despite the added training cost, deploying a smaller model in place of a larger one can have a significant positive impact on compute, inference time \& memory footprint, and thus energy consumption in the long run.
Indicatively, we profiled the inference latency and peak allocated GPU memory of the proposed LLaVA-CKD models with 0.5B and 1.5B LLM backbones, and the TinyLLaVA baseline models with 3B and 7B LLM backbones, using 100 question instances from the MME\textsuperscript{P} dataset and an Nvidia RTX PRO 6000 GPU. This yielded average per-question latency and peak allocated GPU memory of 29 ms, 1.9 GB (0.5B model); 36.5 ms, 3.9 GB (1.5B model); 52.9 ms, 7 GB (3B model); and 78 ms, 15.6 GB (7B model).

\begin{figure*}[tbp]
\centering
\includegraphics[width=0.99\textwidth]{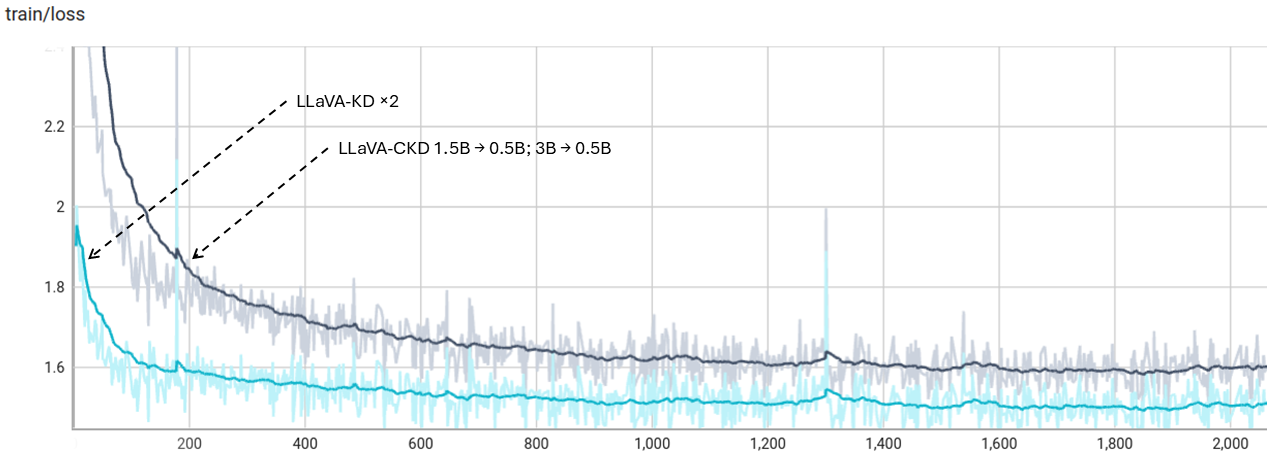}
\caption{Training loss plots (vertical axis: training loss; horizontal axis: number of training steps) of the proposed approach (LLaVA-CKD 1.5B \textrightarrow 0.5B; 3B \textrightarrow 0.5B) vs repeated same-teacher LLaVA-KD (LLaVA-KD $\times 2$).}
\label{fig:ckd_vs_sameta}
\end{figure*}

\begin{figure*}[t]
\begin{center}
\begin{tabular}{cccc}
\hline \\
\includegraphics[align=c, width=0.45\columnwidth]{images/commonsenseReasoning0022.png} &
\includegraphics[align=c, width=0.45\columnwidth]{images/count_000000450303.jpg} &
\includegraphics[align=c, width=0.45\columnwidth]{images/ocr_0020.jpg} &
\includegraphics[align=c, width=0.3\columnwidth]{images/landmark_06483ba2bff074bc.jpg}
\vspace{0.1cm}\\
\parbox{0.45\columnwidth}{\footnotesize Does the animal in the picture usually catch mice?} 
& \parbox{0.45\columnwidth}{\footnotesize Are there six people appear in this image?} 
& \parbox{0.45\columnwidth}{\footnotesize Is the word in the logo cold drinks?}
& \parbox{0.45\columnwidth}{\footnotesize Is this an image of Museo Universitario del Chopo?} 
\vspace{0.1cm}\\
\parbox{0.45\columnwidth}{\footnotesize TLV: \textcolor{red}{No}, LKD: \textcolor{green}{Yes}, CKD: \textcolor{green}{Yes}}
& \parbox{0.45\columnwidth}{\footnotesize TLV: \textcolor{green}{Yes}, LKD: \textcolor{green}{Yes}, CKD: \textcolor{green}{Yes}}
& \parbox{0.45\columnwidth}{\footnotesize TLV: \textcolor{green}{Yes}, LKD: \textcolor{green}{Yes}, CKD: \textcolor{green}{Yes}}
& \parbox{0.45\columnwidth}{\footnotesize TLV: \textcolor{red}{No}, LKD: \textcolor{green}{Yes}, CKD: \textcolor{green}{Yes}}
\vspace{0.2cm}\\
\parbox{0.45\columnwidth}{\footnotesize Is the animal in the picture usually used in search and rescue?} 
& \parbox{0.45\columnwidth}{\footnotesize Are there seven people appear in this image?}
& \parbox{0.45\columnwidth}{\footnotesize Is the word in the logo cold rinks?}
& \parbox{0.45\columnwidth}{\footnotesize Is this an image of Cedar Park, Seattle, Washington?}
\vspace{0.1cm}\\
\parbox{0.45\columnwidth}{\footnotesize  TLV: \textcolor{green}{No}, LKD: \textcolor{green}{No}, CKD: \textcolor{green}{No}}
& \parbox{0.45\columnwidth}{\footnotesize  TLV: \textcolor{red}{Yes}, LKD: \textcolor{red}{Yes}, CKD: \textcolor{green}{No}}
& \parbox{0.45\columnwidth}{\footnotesize  TLV: \textcolor{red}{Yes}, LKD: \textcolor{red}{Yes}, CKD: \textcolor{green}{No}}
& \parbox{0.45\columnwidth}{\footnotesize  TLV: \textcolor{green}{No}, LKD: \textcolor{red}{Yes}, CKD: \textcolor{green}{No}}
\\ \hline \\
\includegraphics[align=c, width=0.45\columnwidth]{images/position_000000395801.jpg} &
\includegraphics[align=c, width=0.3\columnwidth]{images/color_000000530457.jpg} &
\includegraphics[align=c, width=0.45\columnwidth]{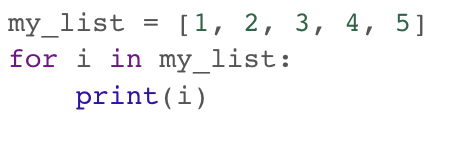} &
\includegraphics[align=c, width=0.45\columnwidth]{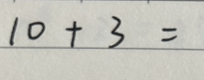}
\vspace{0.1cm}\\
\parbox{0.45\columnwidth}{\footnotesize Is the clock above people?} 
& \parbox{0.45\columnwidth}{\footnotesize Are there any red flowers in the image?} 
& \parbox{0.45\columnwidth}{\footnotesize The image shows a python code. Will the number 3 appear in the output of the code?} 
& \parbox{0.45\columnwidth}{\footnotesize Is the answer to the arithmetic question in the image 13?} 
\vspace{0.1cm}\\
\parbox{0.45\columnwidth}{\footnotesize TLV: \textcolor{green}{Yes}, LKD: \textcolor{red}{No}, CKD: \textcolor{green}{Yes}}
& \parbox{0.45\columnwidth}{\footnotesize TLV: \textcolor{red}{No}, LKD: \textcolor{red}{No}, CKD: \textcolor{green}{Yes}}
& \parbox{0.45\columnwidth}{\footnotesize TLV: \textcolor{green}{Yes}, LKD: \textcolor{green}{Yes}, CKD: \textcolor{green}{Yes}}
& \parbox{0.45\columnwidth}{\footnotesize TLV: \textcolor{green}{Yes}, LKD: \textcolor{red}{No}, CKD: \textcolor{green}{Yes}}
\vspace{0.2cm}\\
\parbox{0.45\columnwidth}{\footnotesize Is the clock under people?} 
& \parbox{0.45\columnwidth}{\footnotesize Are there any green flowers in the image?}
& \parbox{0.45\columnwidth}{\footnotesize The image shows a python code. Will the number 6 appear in the output of the code?}
& \parbox{0.45\columnwidth}{\footnotesize Is the answer to the arithmetic question in the image 12?}
\vspace{0.1cm}\\
\parbox{0.45\columnwidth}{\footnotesize  TLV: \textcolor{red}{Yes}, LKD: \textcolor{red}{Yes}, CKD: \textcolor{green}{No}}
& \parbox{0.45\columnwidth}{\footnotesize  TLV: \textcolor{green}{No}, LKD: \textcolor{green}{No}, CKD: \textcolor{green}{No}}
& \parbox{0.45\columnwidth}{\footnotesize  TLV: \textcolor{green}{No}, LKD: \textcolor{red}{Yes}, CKD: \textcolor{green}{No}}
& \parbox{0.45\columnwidth}{\footnotesize  TLV: \textcolor{green}{No}, LKD: \textcolor{red}{Yes}, CKD: \textcolor{green}{No}}
\\ \hline
\end{tabular}
\end{center}
\caption{Examples from the test set of the MME dataset, across various query categories. The responses of the Qwen2.5-0.5B-based Table \ref{tab:baselines_05b}-reported models created by TinyLLaVA (denoted TLV in this figure), LLaVA-KD 3B \textrightarrow 0.5B (denoted LKD), and the proposed LLaVA-CKD approach (denoted CKD), are shown. Correct / incorrect responses are illustrated in green / red, respectively.}
\label{fig:QuantAnalysis_mme}
\end{figure*}

In order to further document that the improved results of LLaVA-CKD come from distillation using different Teacher/TA networks, rather than just from the increased training cost, we report an ablation where the LLaVA-KD 3B \textrightarrow 0.5B baseline (the best-performing LLaVA-KD setting of Table \ref{tab:baselines_05b}) is repeated twice in a cascade, totaling two training epochs (denoted LLaVA-KD $\times 2$). This matches the total training steps and approximately matches the wall-clock time of the LLaVA-CKD 1.5B \textrightarrow 0.5B; 3B \textrightarrow 0.5B training (in fact, the wall-clock time of this ablation is approximately 6.5\% higher than that of LLaVA-CKD, because the Teacher in both epochs is the larger 3B model). The result of this ablation is reported in the 2nd row of Table \ref{tab:ablTdVsBu}, where we observe that while it surpasses both the single-stage LLaVA-KD (results reported in Table \ref{tab:baselines_05b}) and the computationally heavier top-down distillation, it still lags behind the proposed LLaVA-CKD 1.5B \textrightarrow 0.5B; 3B \textrightarrow 0.5B (average performance 61.8\%, vs. 62.1\% for LLaVA-CKD). Moreover, in Figure \ref{fig:ckd_vs_sameta} we plot the training loss along the training steps for both LLaVA-CKD and LLaVA-KD $\times 2$; in both cases, these curves correspond to the last Distillation finetuning (DFT) step described in Section \ref{ss:LLaVAKD}. We observe that the training loss for LLaVA-KD $\times 2$ drops sharply and almost stabilizes after a few hundred training steps, while for LLaVA-CKD it decreases more smoothly. This, together with the results in Table \ref{tab:ablTdVsBu}, suggests that LLaVA-KD $\times 2$ experiences overtraining to some extent. This is expected, since the Student model during the second repetition of LLaVA-KD $\times 2$ is exposed to the same information as in the first repetition, due to being distilled twice using the same Teacher.   

\subsection{Qualitative Results}
\label{SS:qual_res}

In this section we provide VQA example queries and qualitative results, in order to gain further insight on the proposed approach. Specifically, we show a set of multimodal query examples from four of the evaluation datasets (MME, GQA, VQA\textsuperscript{T} and SQA\textsuperscript{I}), and the corresponding responses of the Table \ref{tab:baselines_05b}-reported models created by the proposed approach (LLaVA-CKD) and our baselines (TinyLLaVA, LLaVA-KD 3B \textrightarrow 0.5B).

\begin{figure*}[!ht]
\begin{center}
\begin{tabular}{cccc}
\hline \\
\includegraphics[align=c, width=0.4\columnwidth]{images/Shelf_n500209.jpg} &
\includegraphics[align=c, width=0.45\columnwidth]{images/Speaker_n77818.jpg} &
\includegraphics[align=c, width=0.45\columnwidth]{images/helicopter_n229548.jpg} &
\includegraphics[align=c, width=0.45\columnwidth]{images/bed_n95313.jpg} 
\vspace{0.1cm}\\ 
\parbox{0.45\columnwidth}{\footnotesize What is in front of the wall that is not short?} 
& \parbox{0.45\columnwidth}{\footnotesize Which kind of device isn't illuminated?} 
& \parbox{0.45\columnwidth}{\footnotesize What is carrying the ship?} 
& \parbox{0.45\columnwidth}{\footnotesize Which color is the bed that the closet is behind of?}
\vspace{0.1cm}\\
\parbox{0.45\columnwidth}{\footnotesize TLV: \textcolor{red}{Book}, LKD: \textcolor{red}{Book}, CKD: \textcolor{green}{Shelf}}
& \parbox{0.45\columnwidth}{\footnotesize TLV: \textcolor{red}{Television}, LKD: \textcolor{red}{Glass}, CKD: \textcolor{green}{Speaker}}
& \parbox{0.45\columnwidth}{\footnotesize TLV: \textcolor{red}{Boat}, LKD: \textcolor{green}{Helicopter}, CKD: \textcolor{green}{Helicopter}}
& \parbox{0.45\columnwidth}{\footnotesize TLV: \textcolor{red}{Blue}, LKD: \textcolor{red}{Blue}, CKD: \textcolor{red}{Blue}}
\vspace{0.2cm}
\\ \hline \\
\includegraphics[align=c, width=0.45\columnwidth]{images/SelectKeybord_a3f1fe4ebd510d4e.jpg} &
\includegraphics[align=c, width=0.45\columnwidth]{images/150825_39c6a622805a6e69.jpg} &
\includegraphics[align=c, width=0.45\columnwidth]{images/pitcher_c47eea828508764e.jpg} &
\includegraphics[align=c, width=0.45\columnwidth]{images/book_657bcc1c27cacad8.jpg} \vspace{0.1cm}\\
\parbox{0.45\columnwidth}{\footnotesize What does the screen say to do?} 
& \parbox{0.45\columnwidth}{\footnotesize What time is on the clock?} 
& \parbox{0.45\columnwidth}{\footnotesize What number is the pitcher wearing?} 
& \parbox{0.45\columnwidth}{\footnotesize What city name is featured on the black book?} 
\vspace{0.1cm}\\
\parbox{0.45\columnwidth}{\footnotesize  TLV: \textcolor{red}{Select your keys}, LKD: \textcolor{red}{Select your key}, CKD: \textcolor{green}{Select your keyboard}}
& \parbox{0.45\columnwidth}{\footnotesize  TLV: \textcolor{green}{15:08:25}, LKD: \textcolor{red}{15:08}, CKD: \textcolor{green}{15:08:25}}
& \parbox{0.45\columnwidth}{\footnotesize  TLV: \textcolor{red}{0}, LKD: \textcolor{green}{39}, CKD: \textcolor{green}{39}}
& \parbox{0.45\columnwidth}{\footnotesize  TLV: \textcolor{red}{New York}, LKD: \textcolor{red}{New York}, CKD: \textcolor{red}{New York}}
\\ \hline \\
\includegraphics[align=c, width=0.45\columnwidth]{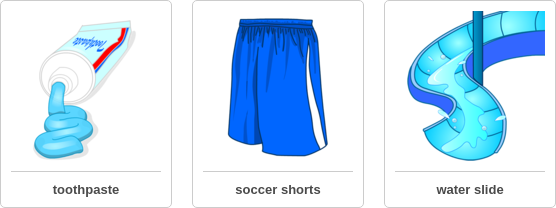} &
\includegraphics[align=c, width=0.45\columnwidth]{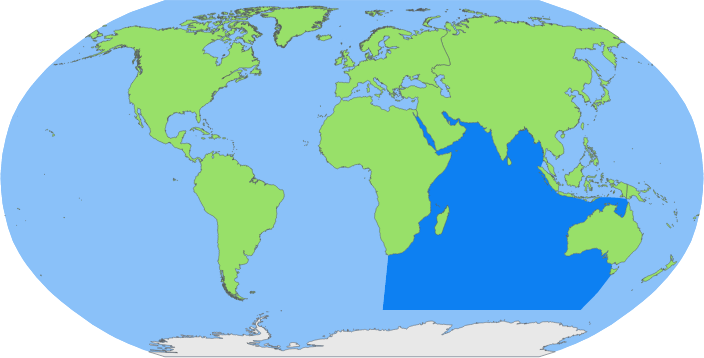} &
\includegraphics[align=c, width=0.45\columnwidth]{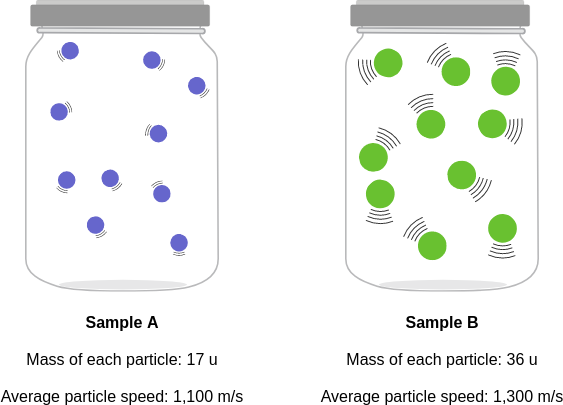} &
\includegraphics[align=c, width=0.45\columnwidth]{images/wettest_1526.png}
\vspace{0.1cm}\\
\parbox{0.45\columnwidth}{\footnotesize Which property do these three objects have in common? A. blue B. hard C. sticky} 
& \parbox{0.45\columnwidth}{\footnotesize Which ocean is highlighted? A. the Indian Ocean B. the Atlantic Ocean C. the Pacific Ocean D. the Southern Ocean.} 
& \parbox{0.45\columnwidth}{\footnotesize The diagrams below show two pure samples of gas in identical closed, rigid containers.  Each colored ball represents one gas particle. Both samples have the same number of particles.Compare the average kinetic energies of the particles in each sample. Which sample has the higher temperature? A. sample A B. sample B C. neither; the samples have the same temperature.}
& \parbox{0.45\columnwidth}{\footnotesize Context: Use the graph to answer the question below. Which month is the wettest on average in Christchurch? A. August B. April C. May.} 
\vspace{0.1cm}\\
\parbox{0.45\columnwidth}{\footnotesize TLV: \textcolor{red}{C}, LKD: \textcolor{green}{A}, CKD: \textcolor{green}{A}}
& \parbox{0.45\columnwidth}{\footnotesize TLV: \textcolor{red}{C}, LKD: \textcolor{green}{A}, CKD: \textcolor{green}{A}}
& \parbox{0.45\columnwidth}{\footnotesize TLV: \textcolor{green}{B}, LKD: \textcolor{red}{A}, CKD: \textcolor{green}{B}}
& \parbox{0.45\columnwidth}{\footnotesize TLV: \textcolor{red}{A}, LKD: \textcolor{red}{A}, CKD: \textcolor{green}{C}}
\\  \hline
\end{tabular}
\end{center}
\caption{Examples from the test set of the GQA, VQA\textsuperscript{T} and SQA\textsuperscript{I} datasets across various query categories are shown in the first, second and third image row, respectively. The responses of the Qwen2.5-0.5B-based Table \ref{tab:baselines_05b}-reported models created by TinyLLaVA (denoted TLV), LLaVA-KD 3B \textrightarrow 0.5B (denoted LKD), and the proposed LLaVA-CKD approach (denoted CKD), are shown. Correct / incorrect responses are illustrated in green / red, respectively.}
\label{fig:QuantAnalysis_gqaVqaSqa}
\end{figure*}

In Figure \ref{fig:QuantAnalysis_mme} we present eight examples from the MME dataset, where each example belongs to a different MME query category.
Specifically, from left to right, in the first row the examples belong to the Commonsense reasoning, Count, OCR and Landmark categories, and in the second row to the Position, Color, Code reasoning and Numerical calculation categories, respectively. For each example image, we present the pair of prompts that are part of the MME dataset and were used for querying the VLMs, as well as the responses of the three models. Similarly, in Figure \ref{fig:QuantAnalysis_gqaVqaSqa} we present examples from the test set of the GQA, VQA\textsuperscript{T} and SQA\textsuperscript{I} datasets across various query categories.

In Figure \ref{fig:QuantAnalysis_mme}, the use of two prompts with contraindicative answers for each image (which is a unique feature of the MME dataset) provides some means of verification that the VLM provided its answer based on the correct evidence, rather than by random guessing. This can be seen in several examples in this figure, where TinyLLaVA and LLaVA-KD provide a wrong answer to one of the questions of each pair; e.g., for the second example in the first row, these methods provide a positive answer to both questions, claiming that the number of people in the image is both six and seven. Contrarily, the LLaVA-CKD model provides more consistent responses.

From both Figures \ref{fig:QuantAnalysis_mme} and \ref{fig:QuantAnalysis_gqaVqaSqa} we observe that the model created by LLaVA-CKD has the potential to provide meaningful responses in a variety of tasks / query categories. For instance, it can perform spatial understanding, distinguish different objects, perform localization, solve simple programming and numerical calculation tasks, and answer questions related to scientific problems. It can also distinguish fine-grained details in images. For instance, in the third example of the first row of Figure \ref{fig:QuantAnalysis_mme} the LLaVA-CKD model can identify that removing letter ``d'' from the word ``drinks'' (2nd query for this example image) changes substantially the meaning of the phrase, and provides the correct answer; while, on the other hand, our two baseline methods fail to identify this subtle difference. Naturally, there are also cases where the LLaVA-CKD model fails; a couple of such failure cases are included in Figure \ref{fig:QuantAnalysis_gqaVqaSqa} (last example in the first two rows of the figure).

\section{Conclusions}
\label{S:Concl}
We introduced a cascaded KD framework to more effectively train a small-scale VLM Student.
Instead of using a one-stage KD (utilizing a single high-performance Teacher), a cascade of KD stages using a TA/Teacher of varying quality is performed to gradually improve the performance of the Student.
The experimental evaluation on seven publicly available benchmarks justified the efficacy of the proposed framework. A \textbf{limitation} of the proposed LLaVA-CKD is that, for a given target Student model, it cannot scale to exceedingly high-quality Teachers: while the cascade gradually reduces the Student-Teacher gap, this reduction would be of diminishing impact if the Student-Teacher gap is too large. \textbf{Future directions} include the application of the proposed LLaVA-CKD to Student-Teacher models that are based on more complex VLM architectures (e.g. mixture-of-experts), and combining more than one TAs of different families in the same cascade chain in order to close the capacity gap between a high-performing Teacher and the Student more effectively. Another future direction is the investigation of the proposed cascaded KD  framework in combination with other base KD techniques, beyond LLaVA-KD.


\begin{IEEEbiography}[{\includegraphics[width=1in,height=1.25in,clip,keepaspectratio]{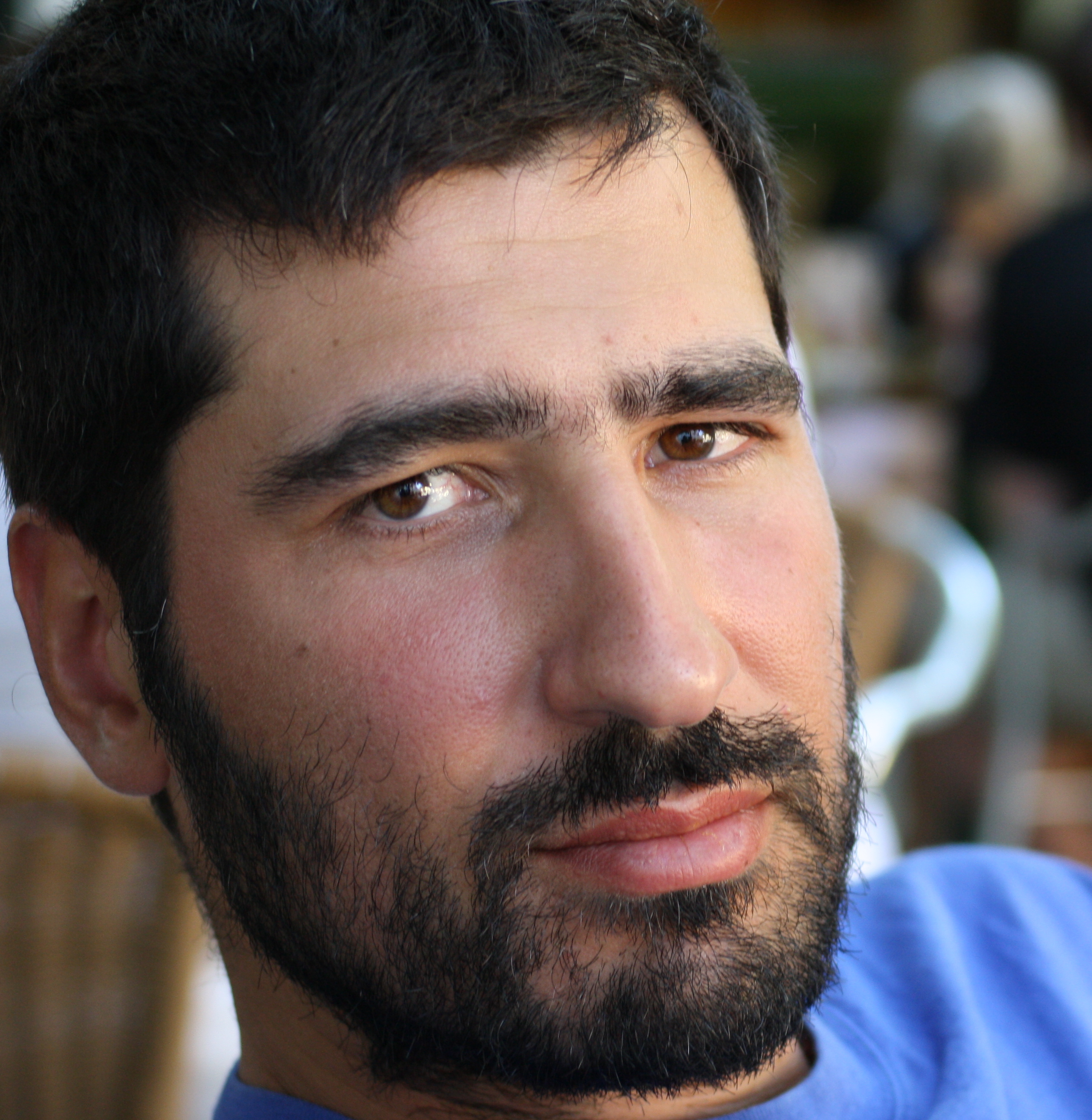}}]
{NIKOLAOS GKALELIS} (Member, IEEE) received the B.Sc. degree in electronic engineering from the Technological Institute of Thessaloniki, Greece, in 2000, the M.Sc. degree in control systems engineering with distinction from the University of Sheffield, U.K., in 2002, the M.Sc. degree in computational engineering from the University of Erlangen-Nuremberg, Germany, in 2004, and the Ph.D. degree in electrical engineering from Imperial College London, U.K., in 2014. He worked in R\&D for Aristotle University of Thessaloniki, Siemens, ECB, and Europol, and as Research Associate and Postdoctoral Researcher at the Centre for Research and Technology Hellas, Information Technologies Institute. He has co-authored 6 journal articles, 1 book chapter, 33 conference papers, and 1 patent. His research interests include deep learning, video analysis, computer vision, and digital signal processing. He serves regularly as a reviewer for international journals and conferences.
\end{IEEEbiography}

\begin{IEEEbiography}[{\includegraphics[width=1in,height=1.25in,clip,keepaspectratio]{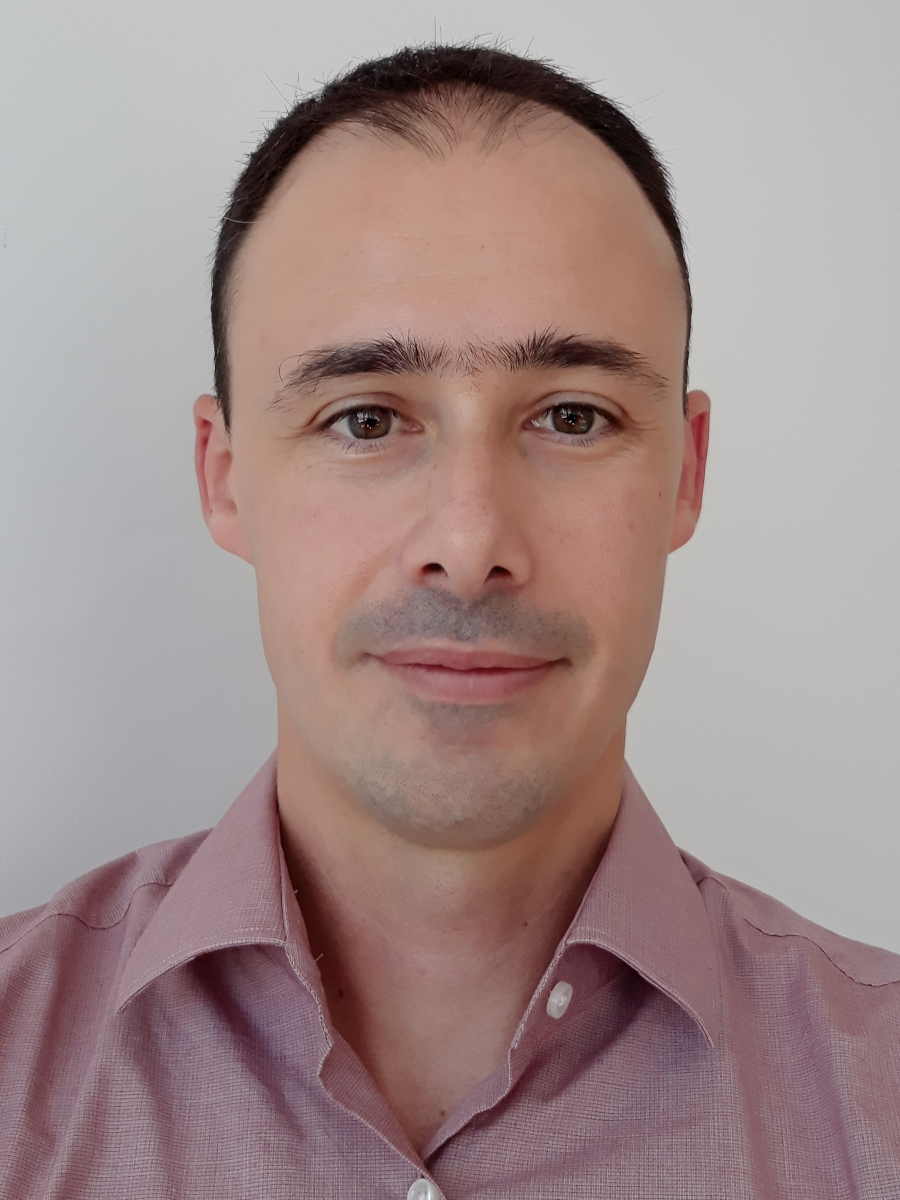}}]
{VASILEIOS MEZARIS} (Senior Member, IEEE) received the B.Sc. and Ph.D. degrees in electrical and computer engineering from the Aristotle University of Thessaloniki, Greece, in 2001 and 2005, respectively. He is currently a Research Director of the Centre for Research and Technology Hellas, Information Technologies Institute. He has coauthored more than 50 journal articles, 20 book chapters, 200 conference papers, and three patents. His research interests include multimedia understanding and artificial intelligence; in particular, image and video analysis and annotation, machine learning and deep learning for multimedia understanding and big data analytics, explainable AI and green AI, multimedia indexing and retrieval, and applications of multimedia understanding and artificial intelligence. He served as Senior Area Editor for IEEE SIGNAL PROCESSING LETTERS, and as Associate Editor for IEEE SIGNAL PROCESSING LETTERS and IEEE TRANSACTIONS ON MULTIMEDIA.
\end{IEEEbiography}


\EOD

\end{document}